**Citation**

```
@article{Atoum2016,
author = {Atoum, Issa and Otoom, Ahmed},
doi = {10.14445/22312803/IJCTT-V31P114},
journal = {International Journal of Computer Trends and Technology (IJCTT)},
keywords = {Clustering,Opinion Mining Tasks,Software Quality-in-use,Topic Models},
number = {2},
pages = {74--83},
title = {{Mining Software Quality from Software Reviews : Research Trends and Open Issues}},
url = {http://www.ijcttjournal.org/2016/Volume31/number-2/IJCTT-V31P114.pdf},
volume = {31},
year = {2016}
}
```

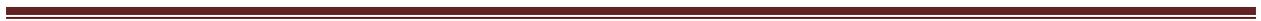

# Mining Software Quality from Software Reviews: Research Trends and Open Issues

Issa Atoum[*1], Ahmed Otoom[2]

[1] Faculty of Computer Information, The World Islamic Sciences & Education University, 11947 Amman, Jordan

[2] Royal Jordanian Air forces,11134 Amman, Jordan

[1] Issa.Atoum@wise.edu.jo

[2] aotoom@rjaf.mil.jo

*Abstract*—Software review text fragments have considerably valuable information about users' experience. It includes a huge set of properties including the software quality. Opinion mining or sentiment analysis is concerned with analyzing textual user judgments. The application of sentiment analysis on software reviews can find a quantitative value that represents software quality. Although many software quality methods are proposed they are considered difficult to customize and many of them are limited. This article investigates the application of opinion mining as an approach to extract software quality properties. We found that the major issues of software reviews mining using sentiment analysis are due to software lifecycle and the diverse users and teams.

## I. INTRODUCTION

The World Wide Web and the social media are an invaluable source of business information. For instance, the software reviews on a website can help users make purchase decisions and enable enterprises to improve their business strategies. Studies showed that online reviews have real economic values [1].The process of extracting information for a decision making from text is referred to as opinion mining or sentiment analysis.

Formally, *"Sentiment analysis or opinion mining refers to the application of natural language processing, computational linguistics, and text analytics to identify and extract subjective information in source materials"* [2, p. 415]. Pang [3] stated that: although many authors use the term *"sentiment analysis"* to refer to classifying reviews as *positive* or *negative*, nowadays it has been taken to mean the computational treatment of opinion, sentiment, and subjectivity in text [3]. Liu [4] identified that the sentiment analysis is more widely used in industry but sentiment analysis and opinion mining are both used in the academia [4]. Both terms are used interchangeably in this article.

Thus, opinion mining is important to organizations and individuals. Organizations can study the products (software) trends over time and respond accordingly. On the other hand, software users often seek advices on software products by reading user reviews found on websites such cnet.com, epinions.com and amazon.com. The software reviews are helpful for users in that it has information about user experience (i.e. Software quality). Garvin [5] identified five views/approaches of quality. The nearest definition to this work is the user based approach definition *"meeting customer needs"*.

To our knowledge little research has been published in the domain of opinion mining over software reviews [6], [7][8], [9]. Mining software reviews can save users time and can help them in software selection process that is time consuming. The most widely used surveys[2], [3], [10] are for products in general and none of them have studied the specialty of a specific review domain. The significance of this article is that it is showed by examples and it details the applicability of sentiment analysis tasks over software quality properties. Further this article identifies major issues to software quality mining using sentiment analysis.

## II. RELATED WORK

Software quality has been studied in many models[11], [12] but [13] found that they are limited. Atoum et *al.* [13] have studied several issues with current software quality models. They showed that studied models are either limited or hard to customize. Atoum et *al.* [14] suggested to build a dataset of software quality-in-use toward solving this problem. They further proposed two frameworks towards solving this problem[6], [15]. A complete model of software prediction were also proposed in [7], [16]. Their frameworks are based on software quality-in-use *keywords* and a built ontology.

Opinion mining can be framed as a text classification task where the categories are polarities (*positive* and *negative*). Text mining has been discussed in topic models[17] and *features* clusters (i.e. grouping) [18]. There are many text classification approaches; Naïve Bayes[19], Support Vector Machines[20], and Maximum Entropy [21].

The semi-supervised learning approaches [20] uses a small set of labelled data and large set of unlabelled data for training. The technique is suitable to take buy in from the user without burdening him with costly labelling for all training data[22], [23], [24].

In the same category a famous family known as topic models are widely used[25][26], [27]. The Latent Semantic (LSA) model [25][26], [27] transforms text to low dimensional matrix and it finds the most common terms that can appear together in the processed text. Wendy et al.[25] applied the LSA in order get the software quality-in-use properties. [28] proposed Probabilistic Latent Semantic Analysis (PLSA) model. The approach aims automatic document indexing based on statistical latent model of counted terms per document.

To our knowledge, little research has been conducted in order to study the sentiment analysis on software quality. Most works considers various products while others are not comprehensive.

### III. PROBLEM DEFINITION

[29] defined opinion mining problem consisting of these components: *topic, opinion holder, sentiment and claim*. [30] defined it the same way but with different components: *opinion holder, subject, aspect, evaluation* where *subject* and *aspect* map to *topic* in Kim model [29], and *evaluation* maps to *claim* and *sentiment*. Probably the most comprehensive definition is given by Liu [2]. An Opinion is defined as $(e_i, a_{ij}, oo_{ijkl}, h_k, t_l)$ where $e_i$ is the name of an *entity*, $a_{ij}$ is an *aspect* of $e_i$, $oo_{ijkl}$ is the orientation of the opinion about *aspect* $a_{ij}$ of *entity* $e_i$, $h_k$ is the *opinion holder*, and $t_l$ is the *time* when the opinion is expressed by $h_k$. if the *entity* is merged with the *aspect* as an opinion target then the definition becomes $(g_i, oo_{ijkl}, h_k, t_l)$ such that $g$ is *topic/entity/properties*. In other words, the $e_j$ and $a_{ij}$ are the *opinion target*. The opinion orientation $oo_{ijkl}$ can be *positive*, *negative* or *neutral*. When an opinion is on the *entity* itself as a whole a special *aspect* called *GENERAL* is used to represent the opinion. Throughout this article Liu [2] definition of opinion mining is adopted.

The Objective of opinion mining, given a set of opinionated reviews $d$, discover all opinion quintuples, then extract *entity, aspects, time, opinion holder* and then assign *sentiment orientation* to *aspects* and group them accordingly. As mentioned earlier, we are concerned with *aspect* extraction, *aspect* assignment orientation and *aspect* grouping tasks.

However, the most important properties of an opinion mining are the *aspects* and *opinions* because the *opinion holder* and the *time* is usually known in software reviews. Furthermore, the concerned *entity* is implied by the software name because software granulates properties at the software level and not as a functional component. Therefore, this article concentrate on *aspect*, *orientation* and *target*.

The example below shows a text fragment of a software review (*AVG antivirus*) extracted from Cnet.com website which was posted on April 20, 2013 by kydna. The numbers indicate the sentence number-sub sentence:

*(1) I've used AVG Free for many years, and have been quite satisfied with it (2-1) until an alert from the software that stated, (2-2)"Resident Shield component not active." (3-1)This means that the program is not updating itself as it should, (3-2) leaving one vulnerable to potential threats. (4-1)Every time I booted up my system, AVG would hang on updating itself, (4-2)chugging and churning for at least 4-5 minutes, (4-3) only to shut down and restart without a current update.*

The opinion according to the previous definition is as follows:
- *Entity*: AVG Free, software, program, system, Resident Shield component.
- *Aspects*: alert, boot, updating chugging and churning, hang.
- *Opinion holder*: Review author (kydna)
- *Time*: April 20, 2013
- *Opinion Orientation*: positive for sentence 1, negative for sentence 2-1,etc.
- Quintuples example:
  (AVG Free, *GENERAL*, *positive*, kydna, April 20, 2013) from sentence (1)

### IV. OPINION MINING TASKS

The main task of sentiment classification is an effective set of *features*. So, given a set of reviews the general opining mining tasks are: identify and extract object *features/entities*, determine the opinion on them, group synonyms of *features,* and finally summarize and present data to users. Below are major research topics and tasks in opinion mining and sentiment analysis grouped in interrelated groups.

#### A. Subjectivity Analysis

Subjectivity classification aims to find if a review sentence is subjective or objective, usually in the presence of an opinion expression in a sentence. A sentence is considered an objective sentence if it has some factual information and is considered subjective if it expresses personal feelings, views, emotions, or beliefs. For example the sentence *"The layers tools need work."* is an objective sentence while the sentence *"that's great antivirus"* is a subjective sentence. However, it is not always easy to detect subjective sentences because sometimes objective sentences can contain opinion, for example the sentence *"To use its best features, you must have the paid version."* is an objective sentence but it indicates a *negative* opinion about the software.

Two classes of subjectivity detection approaches have been proposed; the supervised and the unsupervised learning approaches. In supervised learning approaches, subjectivity classification has been regularly solved as binary classification problem[31]. Pang et al. [32] used min-cut partition based on the assumptions that nearby sentences usually discuss the same topics. [33]used election history to train a SVM on new election posts. [34] proposed an approach to automatically distinguish between subjective and objective segments and between implicit and explicit opinions based on 4 different classes of subjectivity.[35] classified the subjectivity of tweets based on *features* and Twitter clues. In unsupervised learning ,[36] used the presence of subjective expressions extracted using the concept of grade expressions [37]. A gradable expression has a varying strength depending on a standard; for example *the small planet* is larger than the *large house*. [38] used bootstrapping approach to learn two classifiers for subject/objective sentences based on lexical items.

Analyzing the software reviews; sentences are usually short and it is very common to find objective and subjective sentences. For example the sentences *"works for me"* or *"its free"* are common in software reviews. These sentences are objective, but they indicate a *positive* opinion. There are also shorter sentence fragments such as the sentence *"self-updating", "self-regulating", "Ok."* . Consequently, subjectivity analysis is very important to software quality.

### B. *Opinion Lexical Expansion*

To classify a review at the document level, the sentence level or at the *aspect* level, a set of opening words is needed. They are commonly called in literature as sentiment words, opening words, polar words, or opinion bearing words. These words carry the opinion on a specific entity, usually with a *positive* or *negative* polarity. The *positive* sentiment expresses some desired state or qualities whereas the *negative* sentiment words are used to express some undesired sates or qualities. Sentiment words have two types; the base type such as the words *beautiful* and *bad*, and the comparative type such as the words *better*, *best*, *worst*.

The collection of opinion expressions that are used for classification are called the lexicon. A lexicon is the set of opinion words, sentiment phrases and idioms. The lexicon acquisition or expansion is achieved through three techniques: the manual approach, the dictionary-based approach [39]–[43] and the corpus-based approach [37], [44]–[46].The manual approach is not feasible because it is very hard to build a comprehensive lexicon. The dictionary-based approaches use seed opinion words and grow set from an online dictionary like WordNet. The Corpus-based approach discovers additional sentiment words from a domain using general sentiment seeds and adapts general purpose sentiment using a domain corpus. The dictionary-based approach makes it is easy to get words from dictionary but it is domain independent and thus it may not identify the polarity of a word for a specific domain. On the other hand, while corpus-based approach can detect domain specific opinions it is still not easy to build since the same word may have a *positive* or *negative* polarity in the same domain in different contexts[47]. Other lexical expansion approaches are based on dependency parser[48], [49], connotation lexicon [50].

To our knowledge there is no special lexicon for software quality[14], [51]. Thus general lexicon words such as SentiWordNet words could be used. The investigation on a software reviews found that it is uncommon to find one lexicon word with two different orientations. The sentence *"Loads quick, scans quick too!"* is *positive*, while the sentence *"I have always wanted to see a quick disable function rather than clicking on the Resident Shield"* is *negative*.

### C. *Classification*
#### 1) **Classification at the Review Level:**

From information retrieval domain every review can be considered as a single document assuming that each opinionated document expresses an opinion on a single *entity* from a single *holder*. Reviewers have star rating of satisfaction starting by 1 and ending in 5. They can be used for classification (e.g. 1,2 → *negative*, 3→ *neutral*, 4,5→ *positive*) by using any learning algorithm (e.g. Naïve Bayes , SVM or Maximum Entropy.

Other approaches uses the Term-Frequency Inverted-Document-Frequency (TF-IDF) information retrieval model. [52], [53] used review rating regression prediction models on user ratings. Turney proposed unsupervised learning approach [54]. Turney first extracted adjectives and adverbs confirming to a predefined syntactic rules and then estimated the orientation based on Point Mutual Inclusion measure equations from web search engine. Then finally, the average *Sentiment orientation* (SO) is computed for all phrases in the review. The review is classified as recommended if the average *sentiment orientation* is *positive* and not recommended otherwise.

Is it helpful to classify software reviews at the document (review) level? Why? It depends on the needed task. If the task is just user satisfaction, it will be acceptable because sentences are usually short. More practically it can be good to have the classification at the level of review section (e.g. cent *pros*, *cons* or *summary* in cent.com reviews). If user needs to know the underlying topics that are being discussed, then this level will not be helpful.

#### 2) **Classification at the Sentence *Level*:**

To classify a sentence to its sentiments, it is first identified as a subjective or objective sentence. The assumption is that the sentence expresses a single opinion from a single *holder*. Most approaches use supervised learning to learn sentences polarity[31]. [31]

proposed a minimum cuts graph-based approach, assuming that neighboring sentences should have the same subjectivity classification. [55] proposed a lexicon based algorithm to calculate the total orientation by summing the orientation of sentiment words in a sentence. Shein et *al.* [56] proposed to utilize a domain ontology to extract *features* and then they used binary SVM to classify sentences. [57] proposed an unsupervised approach that is based on the average Log-likelihood of words in a sentence. [58] proposed a semi-supervised learning algorithm to learn from a small set of labeled sentences and a large set of unlabeled sentences. [59] identified that conditional sentences has to be taken in their algorithm to deal with different types of if statements. It is noted that sarcastic sentences are not very common in reviews of software reviews.

Is it helpful to classify software reviews at sentence level? Why? Yes if it is linked with underlying topics (*features*). Another problem, many sentences has implicit topics that can be induced at the global sentence level. The sentence *"Stops anything on the internet if there is a problem"*, indicates a *positive* opinion about antivirus protection *feature*. Therefore we should assume that each software sentence is talking about one topic. Consequently, the sentence classification is linked with *feature* classification in order to map topics to sentences.

### D. *Classification at the Feature Level*

The purpose of *aspect (*also called *feature* or *topic)* sentiment classification is to identify the sentiment or the opinion expressed on each *aspect*. The *aspect* sentiment classification methods frequently uses a lexicon , a list of opinion words and phrases to determine the orientation of an *aspect* in a sentence [45], [55]. They first marked opinion words as *positive* or *negative*. Next they handled opinion shifters (valence shifters). Then they aggregated opinion score as the summation of all opinions over the distance between the word and the *aspect*.

Three main approaches are reported in literature; supervised[60] [61] [45] [62] [59], lexicon-based[63], [64] [65] [23], and topic modeling approaches[18], [21], [66], [67]. The supervised approach challenge is how to determine the scope of each sentiment expression over the *aspect* of the interest (i.e. dependency)[60]. Some of these works are discussed in the next section.

Is it helpful to classify software reviews at *feature* level? Why? Yes if it shows user *aspects* and software *features* that makes it good or the software glitches that makes it *bad*.

### E. *Feature Extraction*

Classifying opinion texts at the document or sentence level is insufficient because no opinion targets are defined at that level. Although sentence level classification can give good results, it does not suit compound and complex sentences. [68] showed the emergence need to identify the topic of each sentence. Users need to discover the *aspects* and determine whether the sentiment is *positive* or *negative* on each *aspect*. The purpose of *aspect* sentiment analysis is to determine whether the opinions on different *aspect* are *positive*, *negative* or *neutral*. Given the sentence *"The interface is quite better than the previous version"* and *"Great Antivirus software"* we can say that both of them are *positive* but the second is about the *GENERAL aspect* or the *entity Antivirus* whereas the first is about *interface* of the antivirus. [54, p. 8] clarified that "*the whole is not necessarily the sum of the parts"*. Wilson et *al.* [69] pointed out that the strength of opinions expressed in individual clauses is important as well as pointing out subjective and objective clauses in a sentence. They showed four sentiment levels (*neutral*, low, medium, high).

#### 1) *Explicit Feature Extraction:*

Finding the important aspect of interest for a user is the most important task in sentiment analysis. Feature extraction has been studied in supervised learning approaches [70], [71], frequency based approaches [39], [55], [60], [72]–[75], bootstrapping (from lexicon words or candidate *features)* [48], [49], [76]–[79], and as a topic modeling approaches [18], [21], [66], [67].

In supervised mining [70], [71] proposed to use label sequential rules: The rules that involve a *feature* (called language patterns) are found given that it satisfies predefined support and confidence. Then the sentence segment is matched with language pattern and a *feature* is returned. [75] proposed a supervised based model based on Ku method [80]. The frequency-based approach [55]finds frequent nouns and noun phrases as *aspects*. It also finds infrequent *aspect* exploiting relationships between *aspects* and opinion words [72] [55] [72] [74] [80]. [60] integrated WordNet, and movie reviews to extract frequent *feature* and opinion pairs. [40] refined the frequent noun phrase to consider any noun phrase in sentiment bearings.

Various works [48], [49], [76]–[78] extract domain independent *aspect* and opinion words. Qiu et *al.* [48] [49] double propagation approach is a bootstrapping method based on dependency grammar of [81].[77] extracted *features* that are associated with opinion words and ranked them according to additional patterns. Recently association between *features* and opinions using LSA and likelihood ratio test(LRT) has been employed to find frequent *features* [78]. [79] built iterative learning between *aspects* and opinion words.

Topic modeling approaches based on LDA statistical mixture model have been studied extensively[82][83], however they have the problem of separating features from opinion words [82] [21].

Feature extraction is still an open research area; model-based approaches[54], [55], [72] and statistical models[18], [21], [66] are competing. [84] studied *feature*-learning method completeness from different perspectives such as its ability to identify *features* or opinions words or phrases, ability to reveal intensifiers, ability to classify infrequent *entities*, and ability to classify sentence subjectively. They also studied the application of Conditional Random Fields (CRF) into mining consumer reviews [84].

The feature extraction is the most important part of sentiment analysis task. Without knowing the properties of software quality we can not granulate the overall software quality. For example, the *features fast, load* and *speed* may be mapped to software *efficiency*. The *features work, job and function* can be mapped to *effectiveness* property of software quality. Unlike many methods that use nouns as the baseline of *feature* extraction, in software quality the adjective can still refer to software *feature*. For example in the sentence " this software is fast" , the *keyword fast* my indicate the software speed *feature* (adjective).

2) *Implicit Feature Extraction:*

*Implicit features* can be detected at the global context level and cannot be detected from *features* because usually the *feature* is not found in the sentence. For example the review sentence "*Blocks suspicious and/or alternative sites from opening*", implies that the functionality *feature* is *positive*. Many works takes the *adjective* or *adverb* and sometimes the *verb* as an *implicit feature* indicator[55] [85]. The manual mapping of *implicit features* is difficult. For example, *"This IS a virus. Do not install it or any of its components"*; *virus* here means it is not removing threats or not functioning.

[55] used seed sentiment word to extract infrequent *features* to the opinion word as an indicator of *implicit feature*. [73] applied co-occurrence words between *implicit* and *explicit features* using frequency, PMI variants. [79] extracted *implicit features* by exploiting a function between opinion words and *features*. The current state-of-the-art sequential learning methods are Hidden Markov models HMM [86] [66], and Conditional Random Fields (CRF) [87][88]. [89] used onceClass SVM which trains for *aspects* without any training for non-*aspects*. [90] proposed a supervised learning approach to extract implicit and explicit *aspects*.

Topic modeling is unsupervised learning that assumes each document consists of a mixture of topics and each topic is a probability distribution over words. [91] proposed an *aspect* topic model based on the PLSA.[92] mapped *implicit aspect* expression (sentiment words) to explicit *aspects* using explicit mutual reinforcement relationship between explicit *aspect* and sentiment words.[93] used two phase co-occurrence association rule (explicit *aspects* and sentiment words).

F. *Feature Grouping*

The objective is to group *features* that have the same meaning together. *Aspect* expressions need to be grouped into synonyms *aspect* categories to represent unique *aspects*. For example short words such *AV* may represent an *antivirus*. Liu found that some *aspects* may be referred in many forms; *"call quality"* and *"voice quality"*. Many synonyms are domain dependent [71]; *movie* and *picture* are synonyms in movie reviews but they are not synonyms in camera reviews. The *picture* synonyms refer to *picture* and *movie* refers to *video*.

There are three major approaches for grouping: using semi-supervised learning seeds of *features* and their groups with matching rules [78], [94]–[96], topic models [97], [98] and distributional/relatedness or similarity measures [71], [99]. [94], [95] used semi-supervised learning method to group *aspect* expression into some user specified *aspect* categories using Expectation Maximization algorithm. Zhai et al. [24] used a semi supervised soft-constrained algorithm based on Expectation Maximization(EM) algorithm [100], called soft-constrained EM (SC-EM). [96] extracted domain-independent *features* from reviews and classify them. The [97] algorithm known as DF-LDA, add domain knowledge to topic modeling by incorporating *can-link* and *cannot-link* between *feature* words. Multilevel latent categorization by [98] performs latent semantic analysis to group *aspects* at two levels.

[99] defined several similarity metrics and distances measures using WordNet. It mapped learned *features* into a user-defined taxonomy of the *entity*'s *features* using these measures. The same approach has been used in [71] but both Liu work and Carenini are domain dependent.[78] used the LSA and Likelihood Ratio Test(LRT) as an association between *features* and opinion candidates in order to find real *features* and opinions. [73] grouped explicit synonyms to a predefined most important *features* identified by user using HOWNET nearby synonyms.

It is clear that the candidate *features* in one software category is different the candidate *features* in another category. For example, in antivirus category we can get *features* like *scan, detect, clean*. In internet downloaders category we can get *features* like *speed, kbs, download*. A good grouping approach has to consider the domain specific *features* of each domain.

### G. *Opinion Summarization*

Opinion summarization aims to extract opinions from the text and present them in short form. For document based classification the summary is intuitive where we will get percentage of *positive* opinions versus *negative* opinions. For sentence based classification it is crucial to show users representative sentence for both *positive* and *negative* opinions. The most important summary is the summarization of user opinions on specific *features*. In other words, the Opinion quintuple *aspect* summary to capture the essence of opinion targets (*entities* and *aspects*). It can be used to show percentage of people in different groups based on interest.

There are three major different models to perform summarization of reviews ;1) sentiment match: extract sentences so that the average sentiment of the summary is close to average sentence review of an *entity* [101], 2) sentiment match plus *aspect* coverage(SMAC); a tradeoff *aspect* coverage/sentence *entity* [102], 3) sentiment *aspect* match(SAM); cover important *aspect* with appropriate sentiment [80], [103].[104] proposed an *aspect*-based opinion summarization (or structured summary) to detect Low-quality product review in opinion summarization. [105] used existing online ontology to organize opinions. [106] presented an *aspect* summary layout with a rating for each *aspect*. It identifies *k* interesting *aspects* and cluster head terms into those *aspects*.

What we want to summarize for a software? Why? In software reviews we might be looking for trends in software use over time, software quality, the effect of software enhancement on user usage, the top *k features* that are important to users, the most competitive software to a particular software, etc. These needs could be presented in a graph based approach for users, or managers. If a user gets a graph based view of particular *aspects* of software then he can take a decision in a glance. Software vendor's management teams can know the performance of their product and a marketing strategy or business plans might get updated.

sentiments on topics discussed by the *opinion holder*. Knowing this fact and cross jointing them with the *summary* component we might find a better way to reduce duplication or summarization. For example, if one *aspect* is being repeated in the *summary* component and in the *cons*, it indicates that the *holder* is not happy about that *aspect*. The idea here is to utilize the lengthy *summary* in order to find *implicit* or *explicit aspects* or *entities* that might be difficult to find from the short *cons* or *pros* components of a review.

In our context of software reviews, a possible direction that also has not been studied before is the linkage between the editor review and the *opinion holder* review. The editorial review is usually lengthy and can contain many *features* of the software that it can do, so, why not looking for a way to incorporate this knowledge in opinion mining process in order to find real *features*.

**Annotation schemes limitations:** Annotation is a bottleneck for sentiment analysis (practically software reviews). Many works have verified their models using their own scheme and makes verifying such models reasonably persuasive. Many available annotations do not show details of opinion expressions [55], [74], [107].

Recently [108] proposed a scheme to annotate a corpus of customer reviews by helping annotators using a tool. As a result the annotation contains opinion expression, *opinion target*, a *holder*, *modifier and anaphoric expression*. So results are fine-grain opinion properties that can enhance opinion target extraction and polarity assignment. The previous work showed the importance of an annotation scheme. Furthermore, there is still an immense need for a huge datasets that can be used publically for opinion mining testing similar to the projects of Question Answering challenges (TREC) and text entailments (RTE) projects. Without such type of datasets we believe the many opinion mining techniques will remain questionable unless it is verified by a publically available dataset.

## V. OPEN RESEARCH ISSUES

There are many open research issues in opinion mining as applied on software reviews. We have identified the importance of the below issues:

**Inadequate sentiment analysis models**: currently the most famous model of opinion mining is the general model that represents the *entity*, *aspect*, *opinion*, *opinion holder*, and *time*. Although this model is general and can be used in many domains there might be a researcher need to model the opinion problem for specific domains or review formats. For example, it is generally correct that the *pros* and *cons* of software review documents contain *positive* and *negative*

**Issues software reviews data:** To our knowledge there is no publicly data set that can be used for opinion mining on software reviews[14]. For software that is being developed by a single developer or a small group of users, best practices of software engineering are not usually followed or are ad hoc. Teams are homogenous from the globe and rarely meet face to face. This implies that the software project health status can change dramatically from time to time. As a result at a particular time of software development (or version), the software can get high user satisfaction (many reviews) but at another time it may not get any feedback. A solution of this problem triggers the need of a good opinion mining system that might need to consider the demographics of certain software or to place needed assumptions before mining.

Tight deadlines can force developers to balance quality to time and scope. As a result, the number of users' feedbacks may get down. At this time, users usually start guiding each other to other possible competitive software alternatives.

The software project artefacts are diverse, ranging from the mailing list, forums, source code, change histories, bug reports, etc. So , each of them has an effect of software which means an opinion mining approach might need to consider more than one artefact to get the needed information.

**Noise elimination:** studying reviews, many sentences have grammatical errors and spelling errors. Resolving these errors can enhance dependency parsers. It can enhance *aspect* extraction because sometimes *aspects* are spelled in different ways. Also resolving short text (acronyms) such as the word *gr8* to represent the *great keyword* can enhance opinion orientation. Thus a database of such terms might be helpful. Current spell checkers may need further enhancement to support situations where words are very short or even written in different language. One possible way is to employ a language detector while parsing reviews text to know what the possible written text (by software user).

Another research might be important is the issue of computer generated reviews. Can an opinion mining system detect such type of reviews to eliminate the bias or noise?

**Sentence classification:** while a lot has been done in subjectivity classification, there is a need to filter objective sentences that do not have an opinion whether it is explicit or implied. In fact, this challenge is linked to other challenges such as sarcasm detection, automatic *entity* recognition.

Furthermore, we found more than 40% of a sampled dataset is comparative sentences. Which means users often compare software products to others of the same family. For example, *"Firefox is faster than Internet Explorer."* or *" the previous version is more tidy"*. Therefore, the required *features* and opinion expressions should be extracted from the challenging comparative sentences. In other words, at this point in time of software development the numbers of comparative sentences are very high compared to normal sentences. Given the fact that not all comparative sentences have an opinion special care might be needed to pre-process such reviews during this period. One way might be to link software batch release on the forums or software web sites with such type of reviews.

**Reference resolution**: Reference resolution is important to detect multiple expression /sentences/document referring to the same thing (same referent) that will finally affect the sentiment. For example the sentence *"I bought an IPhone two days ago. It looks very nice. I made many calls"*. Detecting the reference of the article *it* is important or otherwise opinion mining model will lose recall (loose *aspect* opinions. Although there are many researches on reference resolution[109] finding an automatic way to resolve reference and disambiguate word senses is still challenging. It could be helpful for research if taggers can do this job automatically.

**Word sense disambiguation**: Word sense disambiguation[110] is essential for software reviews due the fact that some words are context specific. Sometimes users might review software using asterisks, symbols, or numbers to point out their fulfilment of QinU. In other cases, words might be context specific, thus disambiguation might be required to build good opinion mining systems. For example, in the sentence "I will have to give it *time* for all of the other details", *'time'* here represents time spent by users rather than time spent by the software to do a task (*efficiency*). Furthermore, sometimes sentences are connected with a user story that moves from one *topic* to another.

## VI. CONCLUSIONS

The sentiment analysis tasks and related techniques were studied. We have studied the application of the sentiment analysis on software reviews. We found that there are many issues with sentiment analysis. However the sentiment analysis is promising in detecting software quality. We identified a list of major open issues in sentiment analysis applied on software quality. We conclude that the issues of software quality mining from software reviews are due to the dynamic diverse software lifecycle and the limited software quality datasets.